\documentclass[10pt, a4paper]{article}
\usepackage{lrec}
\usepackage{multibib}
\newcites{languageresource}{Language Resources}
\usepackage{graphicx}
\usepackage{tabularx}
\usepackage{soul}

\usepackage{epstopdf}
\usepackage[utf8]{inputenc}

\usepackage{hyperref}
\usepackage{xstring}

\usepackage{color}

\title{A Survey on Natural Language Processing for Fake News Detection}

\name{Ray Oshikawa$^1$, Jing Qian$^2$, William Yang Wang$^2$}

\address{$^1$College of Arts and Sciences, The University of Tokyo\\ 
$^2$Department of Computer Science, University of California, Santa Barbara\\
         Tokyo, 153-8902 JAPAN / Santa Barbara, CA 93106 USA \\
         ray.oshikawa@gmail.com, \{jing\_qian,william\}@cs.ucsb.edu
         }

\abstract{
Fake news detection is a critical yet challenging problem in Natural Language Processing (NLP). The rapid rise of social networking platforms has not only yielded a vast increase in information accessibility but has also accelerated the spread of fake news. Thus, the effect of fake news has been growing, sometimes extending to the offline world and threatening public safety. 
Given the massive amount of Web content, automatic fake news detection is a practical NLP problem useful to all online content providers, in order to reduce the human time and effort to detect and prevent the spread of fake news.
In this paper, we describe the challenges involved in fake news detection and also describe related tasks.
We systematically review and compare the task formulations, datasets and NLP solutions that have been developed for this task, and also discuss the potentials and limitations of them. Based on our insights, we outline promising research directions, including more fine-grained, detailed, fair, and practical detection models. We also highlight the difference between fake news detection and other related tasks, and the importance of NLP solutions for fake news detection.
 \\ \newline  }

\begin{document}

\maketitleabstract

\section{Introduction}
Automated fake news detection is the task of assessing the truthfulness of claims in news.
This is a new but critical NLP problem because both traditional news media and social media have huge social-political impacts on every individual in the society.
For example, exposure to fake news can cause attitudes of inefficacy, alienation, and cynicism toward certain political candidates \cite{balmas2014fake}.
Fake news even relates to real-world violent events that threaten public safety (e.g., the PizzaGate \cite{kang2016washington}).
Detecting fake news is an important application in the world that NLP can help with, as it also creates broader impacts on how technologies can facilitate the verification of the veracity of claims while educating the general public.

The conventional solution to this task is to ask professionals such as journalists to check claims against evidence based on previously spoken or written facts.
However, it is time-consuming and expensive. For example, PolitiFact\footnote{https://www.politifact.com/} takes three editors to judge whether a piece of news is real or not.

As the Internet community and the speed of the spread of information are growing rapidly, automated fake news detection on internet content has gained interest in the Artificial Intelligence research community.
The goal of automatic fake news detection is to reduce the human time and effort to detect fake news and help us stop spreading it.
The task of fake news detection has been studied from various perspectives with the development in subareas of Computer Science, such as Machine Learning (ML), Data Mining (DM), and NLP.

In this paper, we survey automated fake news detection from the perspective of NLP. 
Broadly speaking, we introduce the technical challenges in fake news detection and how researchers define different tasks and formulate ML solutions to tackle this problem. We discuss the pros and cons, as well as the potential pitfalls and drawbacks of each task. More specifically, we provide an overview of research efforts for fake news detection and a systematic comparison of their task definitions, datasets, model construction, and performances. We also discuss a guideline for future research in this direction.
This paper also includes some other aspects such as social engagement analysis. Our contributions are three-fold:
\begin{itemize}
    \item We provide the first comprehensive review of Natural Language Processing solutions for automatic fake news detection;
    \item We systematically analyze how fake news detection is aligned with existing NLP tasks, and discuss the assumptions and notable issues for different formulations of the problem;
    \item We categorize and summarize available datasets, NLP approaches, and results, providing first-hand experiences and accessible introductions for new researchers interested in this problem. 
\end{itemize}
\begin{table*}[t!]
\begin{center}
\begin{tabular}{|   l    |    l   |    l  |   l    | l |}
\hline \bf Name& \bf Main Input  & \bf Data Size & \bf Label & \bf Annotation \\ \hline
\textsc{liar} & short claim & 12,836 & six-grade & editors, journalists \\
\textsc{fever} & short claim& 185,445 &three-grade& trained annotators\\  
\textsc{buzzfeednews} & FB post & 2,282 & four-grade  &journalists\\
\textsc{buzzface} & FB post & 2,263 &  four-grade &journalists \\ 
\textsc{some-like-it-hoax} & FB post& 15,500 &  hoaxes or non-hoaxes & none\\ 
\textsc{pheme} & Tweet  & 330 &true or false & journalists\\
\textsc{credbank} & Tweet & 60 million & 30-element vector &workers\\
\textsc{fakenewsnet} & article  & 23,921 & fake or real & editors\\
\textsc{bs detector} & article & - & 10 different types & none\\
\hline
\end{tabular}
\caption{A Summary of Various Fake News Detection Related Datasets. \emph{FB: FaceBook.}}
\label{font-table}
\end{center}
\end{table*}

\begin{table*}[t!]
\begin{center}
\begin{tabular}{|   l | l |}
\hline \bf Attributes & \bf Value   \\ \hline
ID of the statement & 11972 \\
Label & True \\
Statement & Building a wall on the U.S.-Mexico border will take literally years. \\
Subject(s) & Immigration\\
Speaker & Rick Perry \\
Speaker's job title & Governor of Texas \\
Party affiliation & Republican     \\
Total Credibility History Counts  & 30, 30, 42, 23, 18 \\
Context & Radio Interview \\
\hline
\end{tabular}
\caption{An Example Entry from \textsc{liar}. The ordered total credibility history counts are \{barely true, false, half true, mostly true, pants on fire\}, note that the history counts only include the history for inaccurate statements.}
\label{LIARtable}
\end{center}
\end{table*}

\section{Related Problems}
\label{RelatedP}
\subsection{Fact-Checking} Fact-checking is the task of assessing the truthfulness of claims made by public figures such as politicians, pundits, etc \cite{vlachos2014fact}. Many researchers do not distinguish fake news detection and fact-checking since both of them are to assess the truthfulness of claims.
Generally, fake news detection usually focuses on news events while fact-checking is broader.
\newcite{thorne2018automated} provides a comprehensive review of this topic.

\subsection{Rumor Detection}
\label{RumorDetection}
There is not a consistent definition of rumor detection. A recent survey \cite{zubiaga2018detection} defines rumor detection as separating personal statements into rumor or non-rumor, where rumor is defined as a statement consisting of unverified pieces of information at the time of posting.
In other words, rumor must contain information that can be verified rather than subjective opinions or feelings.

\subsection{Stance Detection} Stance detection is the task of assessing what side of debate an author is on from text.
It is different from fake news detection in that it is not for veracity but consistency.
Stance detection can be a subtask of fake news detection since it can be applied to searching documents for evidence \cite{ferreira2016emergent}. PHEME, one of the fake-news datasets has tweets related to news, capturing the behavior of users who trust or untrust.
\subsection{Sentiment Analysis} Sentiment analysis is the task of extracting emotions, such as customers' favorable or unfavorable impression of a restaurant.
Different from rumor detection and fake news detection, sentiment analysis is not to do an objective verification of claim but to analyze personal emotions. 

\section{Task Formulations}
\label{Tasks}
In Section \ref{RelatedP}, we compared related problems with fake news detection to define the scope of this survey.
In this survey, The general goal of fake news detection is to identify fake news, defined as the false stories that appear to be news, including rumors judged as information that can be  verified in rumor detection.
Especially, we focus on fake news detection of text content. The input can be text ranging from short statements to entire articles. Inputs are related to which dataset is used (see Section \ref{Datasets}), and additional information such as speakers' identity can be appended. 

There are different types of labeling or scoring strategies for fake news detection. 
In most studies, fake news detection is formulated as a classification or regression problem, but the classification is more frequently used.
\subsection{Classification} The most common way is to formulate the fake news detection as a binary classification problem.
However,  categorizing all the news into two classes (fake or real) is difficult because there are cases where the news is partially real and partially fake.
To address this problem, adding additional classes is common practice.
Mainly, a category for the news which is neither completely real nor completely fake, or, more than two degrees of truthfulness are set as additional classes.
When using these datasets, the expected outputs are multi-class labels, and those labels are learned as independent labels with i.i.d assumptions \cite{rashkin2017truth,wang2017liar}.

One of the conditions for fake news classifiers to achieve good performances is to have sufficient labeled data. 
However, to obtain reliable labels requires a lot of time and labor. Therefore, semi/weakly-supervised and unsupervised methods are proposed \cite{rubin2012identification,bhattacharjee2017active}.

\subsection{Regression} Fake news detection can also be formulated as a regression task, where the output is a numeric score of truthfulness. This approach is used by \newcite{nakashole2014language}.
Usually, evaluation is done by calculating the difference between the predicted scores and the ground truth scores or using Pearson/Spearman Correlations.
However, since the available datasets have discrete ground truth scores, the challenge here is how to convert the discrete labels to numeric scores.

\section{Datasets}\label{Datasets}

A significant challenge for automated fake news detection is the availability and quality of the datasets. 
We categorize public fake-news datasets into three categories: claims, entire articles, and Social Networking Services (SNS) data.
Claims are one or a few sentences including information worth validating (there is a sample in Table \ref{LIARtable}), while entire articles are composed of many sentences related to each other constituting information as the whole. SNS data are similar to claims in length but featured by structured data of accounts and posts, including a lot of non-text data.
\subsection{Claims}
\textsc{PolitiFact}, \textsc {Channel4.com}\footnote{https://www.channel4.com/news/factcheck/}, and \textsc{Snopes}\footnote{https://www.snopes.com/fact-check/} are three sources for manually labeled short claims in news, which is collected and labeled manually. Editors handpicked the claims from a variety of occasions such as debate, campaign, Facebook, Twitter, interviews, ads, etc.
Many datasets are created based on these websites.

\newcite{vlachos2014fact} released the first public fake news detection dataset gathering data from \textsc{PolitiFact} and \textsc {Channel4.com}.
This dataset has 221 statements with the date it was made, the speaker and the URL, and the veracity label of a five-point scale. \textsc{Emergent} \cite{ferreira2016emergent} is the early work of claim-verification dataset too. It is for stance classification in the context of fact-checking, including claim with some documents for or against them. This dataset can improve fact-checking in the condition that some articles related to the claim were given.

Vlachos includes only 221 claims and Emergent includes only 300 claims so that it was impractical to use them for machine learning based assessments.
These days, datasets with many claims are published, which can use as an improved version of the first two.

A recent benchmark dataset for fake news detection is \textsc {liar} \cite{wang2017liar}. 
 This dataset collected data from Politifact as \newcite{vlachos2014fact}, but includes 12,836 real-world short statements, and each statement is labeled with six-grade truthfulness.
 The information about the subjects, party, context, and speakers are also included in this dataset.
 For the datasets from Politifact articles, \newcite{rashkin2017truth} also published large datasets. They collect articles from PunditFact (Politifact's spin-off site) too.
 
 {\textsc fever} \cite{thorne2018fever} is a dataset providing related evidence for fact-checking. In this point, it is similar to \textsc{Emergent}. {\textsc fever} contains 185,445 claims generated from Wikipedia data. Each statement is labeled as \emph{Supported, Refuted, or Not Enough Info}. They also marked which sentences from Wikipedia they use as evidence. {\textsc fever} makes it possible to develop a system that can predict the truthfulness of a claim together with the evidence, even though the type of facts and evidence from Wikipedia may still exhibit some major stylistic differences from those in real-world political campaigns.

\subsection{Entire-Article Datasets}
There are several datasets for fake news detection predicting whether the entire article is true or fake. For example, \textsc{fakenewsnet} \cite{shu2017fake,shu2017exploiting,shu2018fakenewsnet} is an ongoing data collection project for fake news research. It consists of headlines and body texts of fake news articles based on BuzzFeed and PolitiFact. It also collects information about the social engagements of these articles from Twitter.  

\textsc{bs detector}\footnote{https://github.com/bs-detector/bs-detector}\label{BSdetector} is collected from a browser extension named BS Detector, indicating that its labels are the outputs of the BS Detector, not human annotators. 
BS Detector searches all links on a web page at issue for references to unreliable sources by checking against a manually compiled list of unreliable domains. 

\subsection{Posts On Social Networking Services}
There are some datasets for fake news detection focusing on SNS,  but they tend to have a limited set of topics and can be less related to news.

\textsc{buzzfeednews}\footnote{https://github.com/BuzzFeedNews/2016-10-facebook-fact-check} collects 2,282 posts from 9 news agencies on Facebook. Each post is fact\-checked by 5 BuzzFeed journalists. The advantages of this dataset are that the articles are collected from both sides of left-leaning and right-leaning organizations.
There are two enriched versions of \textsc{buzzfeednews}: \newcite{potthast2017stylometric} enriched them by adding data such as the linked articles, and \textsc{buzzface} \cite{santia2018buzzface} extends the BuzzFeed dataset with the
1.6 million comments related to news articles on Facebook.

\textsc{some-like-it-hoax}\footnote{https://github.com/gabll/some-like-it-hoax}  \cite{tacchini2017some}  consists of 15,500 posts from 32 Facebook pages, that is, the public profile of organizations (14 conspiracy and 18 scientific organizations). 
This dataset is labeled based on the identity of the publisher instead of post-level annotations. A potential pitfall of such a dataset is that such kind of labeling strategies can result in machine learning models learning characteristics of each publisher, rather than that of the fake news. 

\textsc{pheme} \cite{zubiaga2016analysing} and \textsc{credbank} \cite{mitra2015credbank} are two Twitter datasets.
\textsc{pheme} contains 330 twitter threads (a series of connected Tweets from one person) of nine newsworthy events, labeled as true or false.
\textsc{credbank} contains 60 million tweets covering 96 days, grouped into 1,049 events with a 30-dimensional vector of truthfulness labels.  Each event was rated on a 5-point Likert scale of truthfulness by 30 human annotators. They concatenate 30 ratings as a vector because they find it difficult to reduce it to a one-dimensional score.

As mentioned above, these datasets were created for verifying the truthfulness of tweets. Thus they are limited to a few topics and can include tweets with no relationship to news. Hence both datasets are not ideal for fake news detection, and they are more frequently used for rumor detection. 

\section{Methods}
\label{Models}
We introduce the methods for fake news detection.
As usual, we first preprocess input texts into suitable forms (\ref{Preprocessing}).
If the dataset has an entire article length, the rhetorical approach can be used as one of the hand-crafted features extraction (\ref{Rhetorical Approach}).
If the dataset has evidence like \textsc{Emergent} or \textsc{FEVER}, we can use methods in \ref{collectevidence} to gather evidence for outputs.

\subsection{Preprocessing}\label{Preprocessing}
Preprocessing usually includes tokenization, stemming, and generalization or weighting words.
To convert tokenized texts into features,  Term Frequency-Inverse Document Frequency (TF-IDF) and Linguistic Inquiry and Word Count (LIWC)\label{LIWC} are frequently used. For word sequences, pre-learned word embedding vectors such as word2vec \cite{mikolov2013efficient} and GloVe \cite{pennington2014glove} are commonly used.
 
When using entire articles as inputs, an additional preprocessing step is to identify the central claims from raw texts. \newcite{thorne2018fever}  rank the sentences using TF-IDF and DrQA system \cite{chen2017reading}.
These operations are closely related to subtasks, such as word embeddings, named entity recognition, disambiguation or coreference resolution.

\subsection{Machine Learning Models}
As mentioned in Section \ref{Tasks}, the majority of existing research uses supervised methods while semi-supervised or unsupervised methods are less commonly used. In this section, we mainly describe classification models with several actual examples.
  
\subsubsection{ Non-Neural Network Models}
Support Vector Machine (SVM) and Naive Bayes Classifier (NBC) are frequently used classification models \cite{conroy2015automatic,khurana2017linguistic,shu2018fakenewsnet}.
These two models differ a lot in structure and both of them are usually used as baseline models.  
Logistic regression (LR) \cite{khurana2017linguistic,bhattacharjee2017active} and decision tree such as Random Forest Classifier (RFC) \cite{hassan2017toward} are also used occasionally.

\subsubsection{Neural Network Models}
\label{NNM}
Recurrent Neural Network (RNN) is very popular in Natural Language Processing, especially Long Short-Term Memory (LSTM), which solves the vanishing gradient problem so that it can capture longer-term dependencies.
In Section \ref{Results}, many models based on LSTM perform high accuracy on \textsc{LIAR} and \textsc{FEVER}.
In addition, \newcite{rashkin2017truth} set up two LSTM models and input text as simple word embeddings to one side and as LIWC feature vectors to the other. In both cases, they were more accurate than NBC and Maximum Entropy(MaxEnt) models, though only slightly.

Convolutional neural networks (CNN) are also widely used since they succeed in many text classification tasks. 
\newcite{wang2017liar}\label{CNN-W} uses a model based on Kim's CNN \cite{kim2014convolutional}, concatenating the max-pooled text representations with the meta-data representation from the bi-directional LSTM.
CNN is also used for extracting features with a variety of meta-data. For example, \newcite{deligiannisdeep} took graph-like data of relationships between news and publishers as input for CNN and assess news with them.

\newcite{karimi2018multi} proposed Multi-source Multi-class Fake news Detection framework (MMFD)\label{MMFD}, in which CNN analyzes local patterns of each text in a claim and LSTM analyze temporal dependencies in
the entire text, then passing
the concatenation of all last hidden outputs through a Fully Connected Network.
This model takes advantage of the characteristics of both models because LSTM works better for long sentences.

Attention mechanisms are often incorporated into neural networks to achieve better performance. \label{kirilin2018exploiting}
\newcite{long2017fake}\label{LSTM-A} used an attention model that incorporates the speaker's name and the statement's topic to attend to features first, then weighted vectors are fed into an LSTM. Doing this increases accuracy by about 3\% (Table \ref{LIARresult}). \newcite{kirilin2018exploiting} used a very similar attention mechanism.
Memory networks, which are a kind of attention-based neural network and also share the idea of attention mechanism, are used by \newcite{pham2018study}. 

\subsection{Rhetorical Approach}
\label{Rhetorical Approach}
Rhetorical Structure Theory (RST)\label{RST},  sometimes combined with the Vector Space Model (VSM), is also used for fake news detection \cite{rubin2015towards,della2018automatic,shu2017exploiting}.
RST is an analytic framework for the coherence of a story. Through defining the semantic role (e.g., a sentence for Circumstance, Evidence, and Purpose) of text units, this framework can systematically identify the essential idea and analyze the characteristics of the input text. Fake news is then identified according to its coherence and structure.

To explain the results by RST, VSM is used to convert news texts into vectors, which are compared to the center of true news and fake news in high-dimensional RST space. Each dimension of the vector space indicates the number of rhetorical relations in the news text. 

\subsection{Collecting Evidence}
\label{collectevidence}
The RTE-based (Recognizing Textual Entailment) \cite{dagan2010recognizing} method is frequently used to gather and to utilize evidence.
RTE is the task of recognizing relationships between sentences.
By gathering sentences that are for or against input from data sources such as news articles using RTE methods, we can predict whether the input is correct or not.
RTE-based models need textual evidence for fact-checking, thus this approach can be used only when the dataset includes evidence, such as \textsc{fever} and Emergent. 

\section{Results \& Observations}
\label{Results}
We compare empirical results on classification datasets via various machine learning models in this section.
 We focused on three datasets: \textsc{liar}, \textsc{fever}, and \textsc{fakenewsnet}.
 We introduced 9 datasets above, but we focus on 3 datasets for looking into the results of experiments on them. It is because others have limited size than newer datasets, limited numbers of experiments, or have the aspect of rumor detection much more than fake news detection.

\subsection{\textsc{LIAR}}
\begin{table}[t!]
\begin{center}
\begin{tabular}{|l|l|l|l|c|}
\hline \bf Author & \bf Meta-data  & \bf Base Model  & \bf Acc. \\ \hline
Wang & & SVMs  &  0.255   \\ 
    &  & CNNs &  0.270  \\ 
   & +Speaker & CNNs  &  0.248  \\ 
  & +All & CNNs  & \textbf{0.274}  \\ \hline

Karimi&  & MMFD  &  0.291  \\ 
 & +All & MMFD&   \textbf{0.348}   \\ \hline

Long& &LSTM+Att &  0.255   \\ 
& +All & LSTM(no Att) &  0.399   \\
& +All & LSTM+Att & \textbf{0.415}    \\ \hline

Kirilin& +All & LSTM &  0.415   \\
& +All+Sp2C & LSTM & \textbf{\underline{0.457}}  \\ \hline
Bhatta-&2-class label& NLP Shallow &  0.921   \\
charjee&  & Deep (CNN)&  0.962  \\ \hline
\end{tabular}
\end{center}
\caption{\label{LIARresult} The Current Results for \textsc{LIAR}. +All means including all meta-data in \textsc{LIAR}. Bhattacharjee convert 6-class labels to 2-class labels.}
\end{table}
Table \ref{LIARresult} shows accuracy of recent studies on \textsc{LIAR}. For the detailed explanations of methods, see Section \ref{Models} As the tendency, LSTM based models 	achieve higher accuracy than CNN based models.
The additional meta-data is also important.
\newcite{karimi2018multi} supplement LIAR by adding the verdict reports written by annotators and raise accuracy by 4\%.  
\newcite{kirilin2018exploiting} improve accuracy by 21\% through replacing the credibility history in LIAR with a larger credibility source (speaker2credit\footnote{https://github.com/akthesis/speaker2credit}).
The two papers also show the attention scores for verdict reports/speaker credit are higher than the statement of claim. 

\subsection{\textsc{fever}}

\begin{table}[t!]
\begin{center}
\begin{tabular}{|l|l|c|}
\hline \bf Author & \bf Model  & \bf Acc. \\ \hline
Thorne& Decomposable Att& \textbf{0.319}  \\ 
&   & 0.509 \\ \hline 
Yin&TWOWINGOS  & {\bf 0.543}   \\ 
&& 0.760  \\\hline 
Hanselowski&LSTM (ESIM-Att) & \underline{ {\bf 0.647} }   \\ 
&& 0.684  \\ \hline 
UNC-NLP&Semantic Matching Network& {\bf 0.640} \\ 
Nie&(LSTM)&  0.680 \\ \hline 
\end{tabular}
\end{center}
\caption{\label{FEVERresult} The Current Results for FEVER. The results in boldface are the accuracy of evidence-collection task.}
\end{table}
Table \ref{FEVERresult} shows accuracy of recent studies on \textsc{FEVER}. 
TWOWINGOS \cite{yin2018twowingos} is the model based on attentive convolution, and \newcite{thorne2018fever} and \newcite{hanselowski2018ukp} also use attention based methods. As \textsc{LIAR}, attention-LSTM has the best score both of verification and evidence-collection task.
The bottom one of the table is the top results of the workshop for FEVER from EMNLP 2018\footnote{http://fever.ai/2018/workshop.html}. This  method selects evidence by conducting semantic matching between each sentence from retrieved pages and the claim.

\subsection{\textsc{fakenewsnet}}
\begin{table}[t!]
\begin{center}
\begin{tabular}{|l|l|l|c|}
\hline \bf Author &Data& \bf Model  & \bf Acc. \\ \hline
Shu&Buzz& RST &  0.610    \\ 
&Feed&LIWC & 0.655    \\ 
&&Castillo & 0.747    \\ 
&&TriFN &  {\bf 0.864} \\ 
Della&&HC-CB-3 &  {\bf 0.856}  \\ 
Deligiannis&&GCN & \textbf{\underline{0.944}}   \\ \hline
Shu&Politi&RST &  0.571    \\  
&Fact&LIWC & 0.637    \\ 
&&Castillo & 0.779   \\
&&TriFN &  {\bf 0.878}   \\  
Deligiannis&&GCN&  {\bf 0.895}      \\
Della&&HC-CB-3   & \textbf{\underline{0.938}} \\ \hline

\end{tabular}
\end{center}
\caption{\label{FNNresult} The Current Results for \textsc{fakenewsnet}. There are two sources of data separately: BuzzFeed and PolitiFact.}
\end{table}
Table \ref{FNNresult} shows accuracy of recent studies on \textsc{fakenewsnet}.
\newcite{shu2017exploiting} achieve over 60\% accuracy by RST and LIWC methods,using hand-selected features such as linguistic or rhetorical features, but both models achieve lower accuracy compared to other methods.
Other methods largely rely on social-engagements data, because \textsc{fakenewsnet} has social engagements of these articles from Twitter.
Castillo, which uses social engagements data only, defeats the model using only textual data (RST, LIWC).
HC-CB-3 sets a threshold of the size of social-engagements data for combining a content-based method and a social-engagements-based method.
GCN takes graph-like data of relationships between news and publishers as input for CNN and assesses news with them.
They achieved very high accuracy by the successful utilization of additional data.

\section{Discussions, \& Recommendations}
\label{Discussions}
\subsection{Datasets and Inputs}
\newcite{rubin2015deception} define nine requirements for fake news detection corpus: 1. Availability of both truthful and deceptive instances; 2. Digital textual format accessibility; 3. Verifiability of ``ground truth''; 4. Homogeneity in lengths; 5. Homogeneity in writing matters; 6. Predefined timeframe; 7. The manner of news delivery; 8. Pragmatic concerns; 9. Consideration for language and culture differences.

As the performances on fake news detection are improved, doing more fine-grained and detailed detection becomes more practical.
We propose new recommendations for a new dataset as the expansion and embodiment of the nine requirements mentioned above, based on the observation of existing dataset and experimental results.

\subsubsection{Sophisticated Index of Truthfulness}
First, news articles or claims might be a mixture of true and false statements, so it is not practical to categorize them totally into true or false.
It is shown by the fact that existing manually fact-checking sites have fine-grained labels such as MOSTLYFALSE, HALFTRUE or Mixture, and annotators find difficult to reduce 30 human annotations to a one-dimensional score while making \textsc{credbank} as mentioned Section \ref{Datasets}
Especially on crowdsourcing, \newcite{roitero2018many} report that
the more classes or the continuous scales seem to lead ordinary people to a similar agreement with expert judges.

Second, as the machine learning and NLP technology have been improved, we already achieve high accuracy on binary classification, especially on claims. In Table \ref{FNNresult} and the bottom 2 case in Table \ref{LIARresult}, the accuracy of 2-class prediction is over 90 \% while 6-way classification only with text is lower then 30\%.
As the next step, we should develop models predicting the veracity of news in more detail than binary assessment. Currently, models on multi-class fake news detection do not concern with the order of labels and just classify. For example, It will be a fatal error if a classifier judges True news as False, but not much in judging True news as Mostly True. However, these two are treated as the same mistake in learning methods so far. It can improve practicality if we can use this distance for learning, so it can be a future issue too.

\subsubsection{Quote claims or articles from various speakers and publishers within the scope of dataset}

Fake news has different nature by its different birth mechanism, for example, some news has the intention to cause harm but other was born only to make fun. There are 7 different types of fake news as defined by Claire Wardle \cite{wardle2017fake}.
\newcite{roitero2018many} shows that satire can be distinguished well from both real and fake news by style analysis, ensuring that the types of fake news are the important factor, so that we should be careful about which types of fake news we will collect.
 
 After defining which types of fake-news the dataset will cover, we should collect data carefully not to label statements solely according to their website source and collect all true or fake news from a certain speaker or publisher.
 \newcite{shu2018fakenewsnet} explore the distribution of publishers who publish
fake news on PolitiFact and GoccipCop, and find that the majority of publishers who published fake news only publish one piece of fake news.
Hence, it is dangerous to assume a publisher as an authentic one because they have not made any mistakes yet. 

In addition, by collecting all true or fake news from a certain situation or publisher, it becomes confounding variables and the task may obtain the aspect of the website classification task.
When collecting data from fact-checking sites, the data has different backgrounds even if the source site is the same; therefore existing datasets frequently based on them.

\subsubsection{Validate Entire Article}
For the claims dataset, there are some sources for manually labeled sentences, such as \textsc{PolitiFact} or {Channel4.com}.
It is easier and cheaper than annotating data from scratch, and additionally, claims are collected grounded, various and natural contexts and labeled with solid analysis.

For the entire article datasets, there is less such websites\footnote{GossipCop may be one of the few but only for celebrity reporting:https://www.gossipcop.com/}. The only human-annotated entire-article dataset is \textsc{Fakenewsnet}, but it is for claims in the article rather than for an entire article, and methods on this dataset emphasize utilization of social-engagements data (Table \ref{FNNresult}).
For those reasons, it is difficult to get human annotation for entire-articles.
But machine annotation has strong assumptions. For example, \textsc{BS detector} only assumes an article's truthfulness by checking domains of links (mentioned in section \ref{Datasets}) and it is not based on its content, so that the machine learning models trained on this dataset are learning the parameters of the BS Detector.
As a future task, we should consider how to evaluate the truthfulness of the entire-article and annotate them. For example, it may be preferable to add truthfulness scores to individual statements.

\subsection{Critiques of Common Methods}
In this section, we analyze different automatic fake news detection solutions and discuss our findings.

First, hand-crafted features were essential in non-neural network approaches but can be replaced by neural networks.

Psycho-linguistic categories and rhetorical features are typical features to extract in fake news detection. After \newcite{mihalcea2009lie} and \newcite{rubin2012identification} find characteristics of the word used and the structure in deceptive languages respectively, \newcite{shu2017exploiting} achieve 60\% accuracy using them(Table \ref{FNNresult}).

However, these hand-crafted features seem to learn something that is more useful and cannot be combined with hand-crafted features. For example, \newcite{rashkin2017truth} shows that adding LIWC did not improve the performance of the LSTM model while non-neural network models are improved largely on their dataset.
There are no existing studies proving the rhetorical features can not be combined with neural network models, but there is a possibility of NN learning something more useful because the RST model achieves lower accuracy compared to other methods (Table \ref{FNNresult}).

Second, the attention mechanism can help improve the performance of fake news detection models.
As a neural network model for Natural Language Processing on fake news detection, LSTM and attention based method such as attention attachments or memory network are often used as mentioned in Section \ref{NNM}
It is because they can analyze long-term and content-transitional information so that they can use the abundant word data of sentences and detect context.
Many research getting high-acurracy in Table \ref{LIARresult},\ref{FEVERresult},\ref{FNNresult} use attention methods or LSTM to learn textual models.

Third, meta-data and additional information can be utilized to improve the robustness and to suppress the noise of a single textual claim or article but should be carefully used. 
Most studies on three datasets improve accuracy by developing a better way to utilize not texts but meta-data including speaker credibility and social engagements' information in section \ref{Results}

However, relying too much on speakers' or publishers' information for judging may cause some problems, such as silencing minorities' voices as Vlachos indicates \cite{graves2018understanding}.
To solve this, he developed a FEVER mentioned above, which includes evidence so that it can be used for claim verification and not only for classification, and the shared task is tackled by many researchers. Considering the top team in the shared tasks used semantic matching networks, the focus on content-based methods may be promoted as intended.
In this point of view, content-based approaches should be developed more in the future, for example, from writing style as \newcite{potthast2017stylometric}.

\section{Conclusion}
\label{Conclusion}
In this survey, we first reveal the importance and definitions of automatic fake news detection.
Then we compare and discuss the most recent benchmark datasets and experimental results of different methods.
Based on our observations, we propose new recommendations for future datasets, and also give the following suggestions for our future fake news detection model:
investigate whether the hand-crafted features can be combined with neural network models,
appropriate usage of non-textual data,
and extending the way of verification with contents.

\bibliographystyle{lrec}
\bibliography{lrec2020W-fakenews}

\begin{thebibliography}{}

\bibitem[\protect\citename{Balmas}2014]{balmas2014fake}
Balmas, M.
\newblock (2014).
\newblock When fake news becomes real: Combined exposure to multiple news
  sources and political attitudes of inefficacy, alienation, and cynicism.
\newblock {\em Communication Research}, 41(3):430--454.

\bibitem[\protect\citename{Bhattacharjee \bgroup et al.\egroup
  }2017]{bhattacharjee2017active}
Bhattacharjee, S.~D., Talukder, A., and Balantrapu, B.~V.
\newblock (2017).
\newblock Active learning based news veracity detection with feature weighting
  and deep-shallow fusion.
\newblock In {\em Big Data (Big Data), 2017 IEEE International Conference on},
  pages 556--565. IEEE.

\bibitem[\protect\citename{Chen \bgroup et al.\egroup }2017]{chen2017reading}
Chen, D., Fisch, A., Weston, J., and Bordes, A.
\newblock (2017).
\newblock Reading wikipedia to answer open-domain questions.
\newblock {\em arXiv preprint arXiv:1704.00051}.

\bibitem[\protect\citename{Conroy \bgroup et al.\egroup
  }2015]{conroy2015automatic}
Conroy, N.~J., Rubin, V.~L., and Chen, Y.
\newblock (2015).
\newblock Automatic deception detection: Methods for finding fake news.
\newblock {\em Proceedings of the Association for Information Science and
  Technology}, 52(1):1--4.

\bibitem[\protect\citename{Dagan \bgroup et al.\egroup
  }2010]{dagan2010recognizing}
Dagan, I., Dolan, B., Magnini, B., and Roth, D.
\newblock (2010).
\newblock Recognizing textual entailment: Rational, evaluation and
  approaches--erratum.
\newblock {\em Natural Language Engineering}, 16(1):105--105.

\bibitem[\protect\citename{Deligiannis \bgroup et al.\egroup
  }2018]{deligiannisdeep}
Deligiannis, N., Do, T.~H., Nguyen, D.~M., and Luo, X.
\newblock (2018).
\newblock Deep learning for geolocating social media users and detecting fake
  news.

\bibitem[\protect\citename{Della~Vedova \bgroup et al.\egroup
  }2018]{della2018automatic}
Della~Vedova, M.~L., Tacchini, E., Moret, S., Ballarin, G., DiPierro, M., and
  de~Alfaro, L.
\newblock (2018).
\newblock Automatic online fake news detection combining content and social
  signals.
\newblock In {\em 2018 22nd Conference of Open Innovations Association
  (FRUCT)}, pages 272--279. IEEE.

\bibitem[\protect\citename{Ferreira and Vlachos}2016]{ferreira2016emergent}
Ferreira, W. and Vlachos, A.
\newblock (2016).
\newblock Emergent: a novel data-set for stance classification.
\newblock In {\em Proceedings of the 2016 conference of the North American
  chapter of the association for computational linguistics: Human language
  technologies}, pages 1163--1168.

\bibitem[\protect\citename{Graves}2018]{graves2018understanding}
Graves, L.
\newblock (2018).
\newblock Understanding the promise and limits of automated fact-checking.
\newblock {\em Factsheet}, 2:2018--02.

\bibitem[\protect\citename{Hanselowski \bgroup et al.\egroup
  }2018]{hanselowski2018ukp}
Hanselowski, A., Zhang, H., Li, Z., Sorokin, D., Schiller, B., Schulz, C., and
  Gurevych, I.
\newblock (2018).
\newblock Ukp-athene: Multi-sentence textual entailment for claim verification.
\newblock {\em arXiv preprint arXiv:1809.01479}.

\bibitem[\protect\citename{Hassan \bgroup et al.\egroup
  }2017]{hassan2017toward}
Hassan, N., Arslan, F., Li, C., and Tremayne, M.
\newblock (2017).
\newblock Toward automated fact-checking: Detecting check-worthy factual claims
  by claimbuster.
\newblock In {\em Proceedings of the 23rd ACM SIGKDD International Conference
  on Knowledge Discovery and Data Mining}, pages 1803--1812. ACM.

\bibitem[\protect\citename{Kang and Goldman}2016]{kang2016washington}
Kang, C. and Goldman, A.
\newblock (2016).
\newblock In washington pizzeria attack, fake news brought real guns.
\newblock {\em the New York Times}.

\bibitem[\protect\citename{Karimi \bgroup et al.\egroup }2018]{karimi2018multi}
Karimi, H., Roy, P., Saba-Sadiya, S., and Tang, J.
\newblock (2018).
\newblock Multi-source multi-class fake news detection.
\newblock In {\em Proceedings of the 27th International Conference on
  Computational Linguistics}, pages 1546--1557.

\bibitem[\protect\citename{Khurana and
  Intelligentie}2017]{khurana2017linguistic}
Khurana, U. and Intelligentie, B. O.~K.
\newblock (2017).
\newblock The linguistic features of fake news headlines and statements.

\bibitem[\protect\citename{Kim}2014]{kim2014convolutional}
Kim, Y.
\newblock (2014).
\newblock Convolutional neural networks for sentence classification.
\newblock {\em arXiv preprint arXiv:1408.5882}.

\bibitem[\protect\citename{Kirilin and Strube}2018]{kirilin2018exploiting}
Kirilin, A. and Strube, M.
\newblock (2018).
\newblock Exploiting a speaker's credibility to detect fake news.
\newblock In {\em Proceedings of Data Science, Journalism \& Media workshop at
  KDD (DSJM'18)}.

\bibitem[\protect\citename{Long \bgroup et al.\egroup }2017]{long2017fake}
Long, Y., Lu, Q., Xiang, R., Li, M., and Huang, C.-R.
\newblock (2017).
\newblock Fake news detection through multi-perspective speaker profiles.
\newblock In {\em Proceedings of the Eighth International Joint Conference on
  Natural Language Processing (Volume 2: Short Papers)}, volume~2, pages
  252--256.

\bibitem[\protect\citename{Mihalcea and Strapparava}2009]{mihalcea2009lie}
Mihalcea, R. and Strapparava, C.
\newblock (2009).
\newblock The lie detector: Explorations in the automatic recognition of
  deceptive language.
\newblock In {\em Proceedings of the ACL-IJCNLP 2009 Conference Short Papers},
  pages 309--312. Association for Computational Linguistics.

\bibitem[\protect\citename{Mikolov \bgroup et al.\egroup
  }2013]{mikolov2013efficient}
Mikolov, T., Chen, K., Corrado, G., and Dean, J.
\newblock (2013).
\newblock Efficient estimation of word representations in vector space.
\newblock {\em arXiv preprint arXiv:1301.3781}.

\bibitem[\protect\citename{Mitra and Gilbert}2015]{mitra2015credbank}
Mitra, T. and Gilbert, E.
\newblock (2015).
\newblock Credbank: A large-scale social media corpus with associated
  credibility annotations.
\newblock In {\em ICWSM}, pages 258--267.

\bibitem[\protect\citename{Nakashole and Mitchell}2014]{nakashole2014language}
Nakashole, N. and Mitchell, T.~M.
\newblock (2014).
\newblock Language-aware truth assessment of fact candidates.
\newblock In {\em Proceedings of the 52nd Annual Meeting of the Association for
  Computational Linguistics (Volume 1: Long Papers)}, volume~1, pages
  1009--1019.

\bibitem[\protect\citename{Pennington \bgroup et al.\egroup
  }2014]{pennington2014glove}
Pennington, J., Socher, R., and Manning, C.
\newblock (2014).
\newblock Glove: Global vectors for word representation.
\newblock In {\em Proceedings of the 2014 conference on empirical methods in
  natural language processing (EMNLP)}, pages 1532--1543.

\bibitem[\protect\citename{Pham}2018]{pham2018study}
Pham, T.~T.
\newblock (2018).
\newblock A study on deep learning for fake news detection.

\bibitem[\protect\citename{Potthast \bgroup et al.\egroup
  }2017]{potthast2017stylometric}
Potthast, M., Kiesel, J., Reinartz, K., Bevendorff, J., and Stein, B.
\newblock (2017).
\newblock A stylometric inquiry into hyperpartisan and fake news.
\newblock {\em arXiv preprint arXiv:1702.05638}.

\bibitem[\protect\citename{Rashkin \bgroup et al.\egroup
  }2017]{rashkin2017truth}
Rashkin, H., Choi, E., Jang, J.~Y., Volkova, S., and Choi, Y.
\newblock (2017).
\newblock Truth of varying shades: Analyzing language in fake news and
  political fact-checking.
\newblock In {\em Proceedings of the 2017 Conference on Empirical Methods in
  Natural Language Processing}, pages 2931--2937.

\bibitem[\protect\citename{Roitero \bgroup et al.\egroup
  }2018]{roitero2018many}
Roitero, K., Demartini, G., Mizzaro, S., and Spina, D.
\newblock (2018).
\newblock How many truth levels? six? one hundred? even more? validating
  truthfulness of statements via crowdsourcing.
\newblock In {\em Proceedings of the 2nd International Workshop on Rumours and
  Deception in Social Media}.

\bibitem[\protect\citename{Rubin and Vashchilko}2012]{rubin2012identification}
Rubin, V.~L. and Vashchilko, T.
\newblock (2012).
\newblock Identification of truth and deception in text: Application of vector
  space model to rhetorical structure theory.
\newblock In {\em Proceedings of the Workshop on Computational Approaches to
  Deception Detection}, pages 97--106. Association for Computational
  Linguistics.

\bibitem[\protect\citename{Rubin \bgroup et al.\egroup
  }2015a]{rubin2015deception}
Rubin, V.~L., Chen, Y., and Conroy, N.~J.
\newblock (2015a).
\newblock Deception detection for news: three types of fakes.
\newblock In {\em Proceedings of the 78th ASIS\&T Annual Meeting: Information
  Science with Impact: Research in and for the Community}, page~83. American
  Society for Information Science.

\bibitem[\protect\citename{Rubin \bgroup et al.\egroup
  }2015b]{rubin2015towards}
Rubin, V.~L., Conroy, N.~J., and Chen, Y.
\newblock (2015b).
\newblock Towards news verification: Deception detection methods for news
  discourse.
\newblock In {\em Proceedings of the Hawaii International Conference on System
  Sciences (HICSS48) Symposium on Rapid Screening Technologies, Deception
  Detection and Credibility Assessment Symposium, January}, pages 5--8.

\bibitem[\protect\citename{Santia and Williams}2018]{santia2018buzzface}
Santia, G.~C. and Williams, J.~R.
\newblock (2018).
\newblock Buzzface: A news veracity dataset with facebook user commentary and
  egos.
\newblock In {\em ICWSM}, pages 531--540.

\bibitem[\protect\citename{Shu \bgroup et al.\egroup }2017a]{shu2017fake}
Shu, K., Sliva, A., Wang, S., Tang, J., and Liu, H.
\newblock (2017a).
\newblock Fake news detection on social media: A data mining perspective.
\newblock {\em ACM SIGKDD Explorations Newsletter}, 19(1):22--36.

\bibitem[\protect\citename{Shu \bgroup et al.\egroup }2017b]{shu2017exploiting}
Shu, K., Wang, S., and Liu, H.
\newblock (2017b).
\newblock Exploiting tri-relationship for fake news detection.
\newblock {\em arXiv preprint arXiv:1712.07709}.

\bibitem[\protect\citename{Shu \bgroup et al.\egroup }2018]{shu2018fakenewsnet}
Shu, K., Mahudeswaran, D., Wang, S., Lee, D., and Liu, H.
\newblock (2018).
\newblock Fakenewsnet: A data repository with news content, social context and
  dynamic information for studying fake news on social media.
\newblock {\em arXiv preprint arXiv:1809.01286}.

\bibitem[\protect\citename{Tacchini \bgroup et al.\egroup
  }2017]{tacchini2017some}
Tacchini, E., Ballarin, G., Della~Vedova, M.~L., Moret, S., and de~Alfaro, L.
\newblock (2017).
\newblock Some like it hoax: Automated fake news detection in social networks.
\newblock {\em arXiv preprint arXiv:1704.07506}.

\bibitem[\protect\citename{Thorne and Vlachos}2018]{thorne2018automated}
Thorne, J. and Vlachos, A.
\newblock (2018).
\newblock Automated fact checking: Task formulations, methods and future
  directions.
\newblock {\em COLING}.

\bibitem[\protect\citename{Thorne \bgroup et al.\egroup }2018]{thorne2018fever}
Thorne, J., Vlachos, A., Christodoulopoulos, C., and Mittal, A.
\newblock (2018).
\newblock Fever: a large-scale dataset for fact extraction and verification.
\newblock {\em arXiv preprint arXiv:1803.05355}.

\bibitem[\protect\citename{Vlachos and Riedel}2014]{vlachos2014fact}
Vlachos, A. and Riedel, S.
\newblock (2014).
\newblock Fact checking: Task definition and dataset construction.
\newblock In {\em Proceedings of the ACL 2014 Workshop on Language Technologies
  and Computational Social Science}, pages 18--22.

\bibitem[\protect\citename{Wang}2017]{wang2017liar}
Wang, W.~Y.
\newblock (2017).
\newblock " liar, liar pants on fire": A new benchmark dataset for fake news
  detection.
\newblock {\em arXiv preprint arXiv:1705.00648}.

\bibitem[\protect\citename{Wardle}2017]{wardle2017fake}
Wardle, C.
\newblock (2017).
\newblock Fake news. it's complicated.
\newblock {\em First Draft News}.

\bibitem[\protect\citename{Yin and Roth}2018]{yin2018twowingos}
Yin, W. and Roth, D.
\newblock (2018).
\newblock Twowingos: A two-wing optimization strategy for evidential claim
  verification.
\newblock {\em arXiv preprint arXiv:1808.03465}.

\bibitem[\protect\citename{Zubiaga \bgroup et al.\egroup
  }2016]{zubiaga2016analysing}
Zubiaga, A., Liakata, M., Procter, R., Hoi, G. W.~S., and Tolmie, P.
\newblock (2016).
\newblock Analysing how people orient to and spread rumours in social media by
  looking at conversational threads.
\newblock {\em PloS one}, 11(3):e0150989.

\bibitem[\protect\citename{Zubiaga \bgroup et al.\egroup
  }2018]{zubiaga2018detection}
Zubiaga, A., Aker, A., Bontcheva, K., Liakata, M., and Procter, R.
\newblock (2018).
\newblock Detection and resolution of rumours in social media: A survey.
\newblock {\em ACM Computing Surveys (CSUR)}, 51(2):32.

\end{thebibliography}

\end{document}